\documentclass[10pt,twocolumn,letterpaper]{article}

\usepackage{iccv}
\usepackage{times}
\usepackage{epsfig}
\usepackage{graphicx}
\usepackage{amsmath}
\usepackage{amssymb}

\usepackage{custom_notations}
\usepackage{multirow}
\usepackage{subfigure}
\usepackage{comment}
\usepackage{float}
\usepackage[usenames]{color}
\usepackage{tabularx}
\usepackage{diagbox}

\usepackage[breaklinks=true,
            bookmarks=false,
            colorlinks,
            linkcolor=red,
            anchorcolor=blue,
            citecolor=green
            ]{hyperref}

\iccvfinalcopy 

\def\iccvPaperID{****} 

\begin{document}

\title{TOP: Backdoor Detection in Neural Networks via Transferability of Perturbation}

\author{Todd Huster, Emmanuel Ekwedike\\
Perspecta Labs\\
Basking Ridge, NJ, USA\\
{\tt\small \{thuster, emmanuel.ekwedike\}@perspectalabs.com}
}

\maketitle
\ificcvfinal\thispagestyle{empty}\fi

\begin{abstract}
 \noindent Deep neural networks (DNNs) are vulnerable to ``backdoor" poisoning attacks, in which an adversary implants a secret trigger into an otherwise normally functioning model. Detection of backdoors in trained models without access to the training data or example triggers is an important open problem. In this paper, we identify an interesting property of these models: adversarial perturbations transfer from image to image more readily in poisoned models than in clean models. This holds for a variety of model and trigger types, including triggers that are not linearly separable from clean data. We use this feature to detect poisoned models in the TrojAI benchmark, as well as additional models.
\end{abstract}

\section{Introduction}
\noindent Deep Neural Networks (DNNs) have achieved success in many mission-critical tasks, such as malware detection, face recognition, autonomous driving, medical diagnosis, etc. \cite{Redmon2016LookOnce, Tobiyama2016MalDetect, Parkhi2015DeepFR, bojarski2016end,Bakator2018DLMD}. However, it has been shown that DNNs are vulnerable to backdoor poisoning attacks, in which an adversary implants a secret trigger into an otherwise normally functioning model \cite{Chen2017_Targeted_Backdoor, Liu2018TrojaningAO, NeuralTrojansSurvey2020, li2021backdoor}. In backdoor attacks, an adversary embeds a hidden backdoor into DNNs, such that the compromised DNN model performs well on benign samples and misclassifies when the hidden backdoor is activated by an adversary-defined trigger. These backdoor attacks do not seek to degrade classification accuracy, but rather to control the behavior of the classifier by adding the trigger to data \cite{xiang2020detection, gao2020backdoor}. 

Detecting the presence of a backdoor in a trained model without access to examples of the trigger is an important open problem \cite{chen2018detecting, NeuralCleanse2019, xiang2020detection}.
A number of works have sought to detect poisoning by examining the internal network activations during training or at run time \cite{liu2017neural,Ma2019NICDA, zeng2020deepsweep}. However, a neural network training supply chain is complex \cite{Jansen2020DDLBenchTA}, so assuming access to all training data is often unrealistic. Furthermore, monitoring a model for run time anomalies is expensive and likely too late \cite{wang2020practical}. Therefore, it is important to be able to identify poisoning by inspecting the model itself. 

To this end, we identify a feature of poisoned models that is detectable without access to any examples of the trigger: the transferability of their adversarial perturbations (TOP) from one image to another.
Our solution does not assume access to training data or specific information about the trojan trigger.
It requires only the pre-trained model and a small set of benign inputs representative of the domain. 
\newline\noindent\textbf{Main Contributions.} We identify TOP, a novel property of poisoned neural networks, and develop a reliable method for using this property to detect poisoning. 
We validate the TOP method in a variety of settings, including a standard benchmark for backdoor detection. Finally, we show that this method is effective without extensive tuning or training.

\section{Preliminaries}\label{section:preliminaries}
\noindent In this section, we briefly introduce our main notations and describe our threat model.
\subsection{Notation}
\noindent Let $x \in \RR^d$ be a $d$-dimensional image from some naturally occurring distribution $p(x)$.
Let $class: \RR^d \to C=\{c_1,...,c_m\}$ represent a class mapping of an arbitrary image. For simplicity, we assume that $class$ exists and is deterministic. 
We use $f: \RR^d \to \cY$ to denote a neural network, where $\cY$ is a probability simplex over classes $C$.
We define a function $\delta(f,x)$ representing an untargeted adversarial ``evasion" attack \big(e.g., PGD \cite{Madry2018TowardsDL}\big) on neural network $f$ for input $x$ over some attack domain $\Delta$ and loss function $\cL$:
\begin{equation}\label{eq:delta}
\delta(f,x) = \argmax_{\delta \in \Delta}\cL\Big(f(x), f(x+\delta)\Big)
\end{equation}
For the loss function $\cL$, we use cross entropy between the adversarial prediction and the original class prediction, equivalent to:
\begin{equation}
\cL(y, \hat{y}) = -\log \hat{y} \cdot\one\big\{\argmax(y)\big\},
\end{equation}
where $\one$ is the indicator function. This is the standard loss function used in untargeted evasion attacks \cite{ding2019advertorch}.

\subsection{Threat model}
\noindent We briefly describe the assumptions of the attacker and the defender.

\noindent\textbf{Setting.} Our threat model considers a user who wishes to utilize a DNN trained in an untrusted environment. For example, the model could be trained by a third party, in a cloud computing environment, or on a system infected with malware. 
The user knows the basic DNN functionality (e.g., the nature of the expected input data, what classes it is trained on), and has some validation data to test performance on.
Since the user has the ability to examine the validation data, it is presumed to be clean and correctly labeled. 
The user checks the accuracy of the trained model on the validation dataset and only deploys the DNN if it has satisfactory validation accuracy.

\noindent\textbf{Attacker.} We make the following assumptions about the resources and goals of the attacker. The attacker has full control over the training procedure and the training dataset, but does not have access to the data used for validation. 
The attacker can train the DNN on an arbitrary dataset, including any number of poisoned training inputs, and can manipulate the training process in arbitrary ways. 

The attacker's goal is to produce a maliciously backdoored model that satisfies two properties. First, it must pass the user's validation threshold. 
Second, applying a trigger function $T: \RR^d \to \RR^d$ to a normal image $x$ must change the model's behavior in a prescribed way. 
We make the following assumptions about trigger functions. First, a triggered image $T(x)$ is unlikely to occur naturally. Second, trigger functions rarely change the class \big(i.e., $\PP\big(class\big(T(x)\big) = class(x)\big) \approx 1$ \big). Lastly, $T(x)$ and $x$ have a small ``magnitude'' difference, with magnitude related in some way to human perception. We discuss different approaches for quantifying magnitude in Section \ref{section:reverse_eng_trigger}.
For the purposes of this paper, we assume that the presence of the trigger should change the backdoored model's predictions from some set of source classes $C_s \subseteq C$ to a target class $c_t \in C$. 

\noindent\textbf{Defender.} We make the following fundamental assumption about the resources available to the defender. The defender does not have access to the backdoor training data or process, since training takes place in an untrusted environment. The defender only has access to the trained DNN and a small validation dataset. We presume the validation set to be clean and correctly labeled, as it can be manually verified or spot-checked by the defender. The goal of the defender is to determine whether or not a given DNN has been infected by a backdoor.

\section{Related Work}
\noindent We briefly discuss several existing backdoor attacks and detection techniques besides those discussed in the introduction.

\noindent\textbf{Backdoor (Trojan) Attacks.} Neural network vulnerabilities have been exploited by backdoor attacks \cite{gu2019badnets,Turner2018CleanLabelBA}. BadNets by Gu \etal \cite{gu2019badnets} first explored the vulnerabilities of the DNN by injecting backdoors into a neural network via dataset poisoning. While BadNets require clearly mislabeled in the training set, other proposed attacks are more covert, performing poisoning without obvious mislabeling \cite{Turner2018CleanLabelBA, saha2019hidden, Liu2018TrojaningAO}.
Moreover,  Liu \etal \cite{Liu2018TrojaningAO} shows a backdoor attack on DNN is also possible without access to the original clean training data and without the need to compromise the original training process.  A comprehensive survey of backdoor attacks can be found in \cite{NeuralTrojansSurvey2020,li2021backdoor}.

\noindent\textbf{Backdoor Detection.} Several backdoor detection techniques have been proposed in the literature \cite{NeuralCleanse2019, guo2019tabor, gao2020strip, zeng2020deepsweep, liu2018finepruning, chen2019deepInspect}. Existing backdoor detection techniques can be broadly categorized as either detecting malicious inputs at runtime \cite{liu2017neural,Ma2019NICDA, zeng2020deepsweep, chou2020sentinet,gao2020strip} or scanning models to determine if they have backdoors \cite{chen2018detecting, guo2019tabor, NeuralCleanse2019, wang2020practical, sun2020poisoned, shen2021backdoor, bajcsy2021baseline}. 
The former, detecting malicious inputs at runtime, often relies on having access to a poisoned input to decide the malicious identity of a model.  Both SentiNet \cite{chou2020sentinet} and STRIP \cite{gao2020strip}  undertake a run-time detection of backdoor by examining inputs of an actively deployed model.  Similarly, Chen \etal \cite{chen2018detecting} proposed an Activation Clustering methodology for detecting and removing backdoors. While Activation Clustering demonstrates the effectiveness in detecting trojan backdoors, it assumes access to the training data.  

We focus on the case where one does not require poisoned input or the training data to decide the malicious identity of the model. 
Neural Cleanse by Wang \etal \cite{NeuralCleanse2019} proposes an optimization technique for detecting and reverse engineering hidden triggers embedded inside deep neural networks for each class. Similarly, TABOR by  Guo \etal \cite{guo2019tabor}  formalizes the detection of trojan backdoors as an optimization problem and identifies a set of candidate triggers by resolving this optimization problem. Both Neural Cleanse and TABOR attempt to reconstruct the backdoor and require solving custom-designed optimization problems.

Our technique is also inspired by the work of Sun \etal \cite{sun2020poisoned}, which showed that backdoor attacks create poisoned classifiers that can be easily attacked even without knowledge of the original backdoor. We build upon this work by identifying an inherent property of poisoned neural networks that is detectable without access to any examples of the trigger. This inherent property of poisoned neural networks is based on the transferability of adversarial perturbations.
\section{Transferability}\label{section:transferability}
\begin{figure}[!htbp]
  \centering
    \includegraphics[width=.45\textwidth]{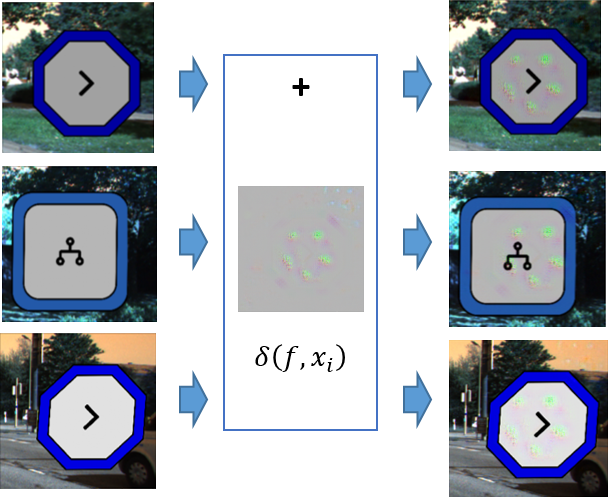} 
    \caption{Adversarial perturbation from image $x_i$ added to additional images.}\label{fig:adversarial_perturbation} 
\end{figure}

\noindent Our motivating observation is that adversarial perturbations readily transfer from image to image in poisoned models, whereas clean models are more likely to resist such a transfer. We define a transfer attack from image $x_i$ to image $x_j$ as follows:
\begin{equation}
\tilde{x}_j = x_j + \delta(f,x_i).
\end{equation}
Figure \ref{fig:adversarial_perturbation} shows an adversarial perturbation from one image being added to three different images. 
We use two metrics to quantify the characteristics of transfer attacks for a particular model. 
Given a set of $n$ sample images $\{x_1,\dots,x_n \}$, fool rate (FR) is the proportion of images whose class predictions change under a transfer attack: 
\begin{equation}
\text{FR} = \frac{1}{n^2}\sum_{i=1}^n\sum_{j=1}^n \one \Big\{f(x_i)\neq f\big(x_i +\delta(f,x_j)\big) \Big\}.
\end{equation}
Fool concentration (FC) is concerned which class most predictions get changed to. Intuitively, reverse engineering the trigger of a poisoned model will lead to an out-sized proportion of images changing to the (unknown) target class. 
FC detects this behavior by measuring how often perturbations change predictions to a particular class. 
Formally, we define FC as follows:
\begin{equation}
\text{FC} = \max_{k \in \{1,2,\ldots, m\}}\frac{1}{n^2}\sum_{i=1}^n\sum_{j=1}^n \one\big\{\Xi_{ijk}\big\},
\end{equation}
\begin{equation}
\text{where}  \,\,\Xi_{ijk}\equiv \Big(f\big(x_i +\delta(f, x_j)\big)= c_k\Big) \wedge \Big(f(x_i)\neq c_k\Big) .\nonumber    
\end{equation}


\begin{figure}[!htbp]
  \centering
    \includegraphics[width=.45\textwidth]{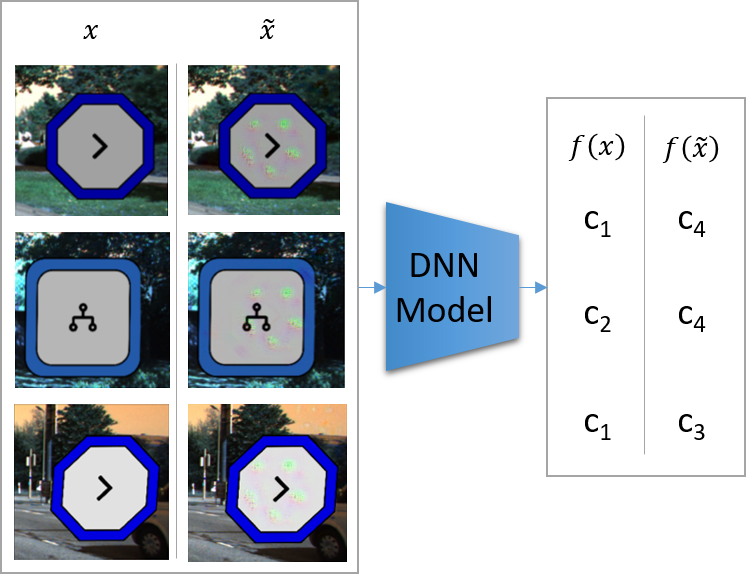} 
    \caption{Clean and perturbed images and the model's class predictions. Since all images change class, $FR=1$.  Since two images change to $c_4$, $FC=2/3$.}\label{fig:transfer_metrics} 
\end{figure}
Note that both of these metrics are functions of a particular model and a particular attack. 
Our observation is that both FR and FC tend to be larger for a poisoned model than a clean one over a broad class of attacks.  
This makes intuitive sense for many common triggers, such as the ``checkerboard'' pattern from Gu \etal \cite{gu2019badnets} or a watermark trigger from Liu \etal \cite{Liu2018TrojaningAO}. Models poisoned with these triggers essentially have a universal adversarial perturbation built into them \cite{UAP2017}; so as long as $\delta(f,x_i)$ successfully reverse engineers this perturbation, a transfer attack should be successful. More generally, it is reasonable to assume that this behavior exists for any trigger that is linearly separable from clean data. Suppose there exists a hyperplane defined by $W \in \mathbb{R}^d$, $b \in \mathbb{R}$ such that
\begin{equation}
\PP\big(Wx+b>0\big) \approx 0, \, \text{and} \,\,  \PP\big(W\cdot T(x)+b>0\big) \approx 1.
\end{equation}
Such a hyperplane is easy for a neural network to learn during training, and the vector $W$, projected onto the attack domain $\Delta$, will be an effective attack for any image outside of the target class. Surprisingly, though, this phenomenon holds even when triggered data is not linearly separable from clean data. We designed two non-linear triggers and trained poisoned models with them:

\noindent\textbf{3-Pixel Flip Trigger.} This trigger rotates 3 specific pixels $x^{(1)},x^{(2)},x^{(3)}$ around their means.  We set $x^{(i)} := 2\mu^{(i)} - x^{(i)}$ for $i \in \{1,2,3\}$ where $\mu^{(i)}$ is the mean of $x^{(i)}$.

\noindent\textbf{CDF Flip Trigger.} This trigger flips a single pixel $x^{(1)}$, while preserving its per-class marginal distribution. This ensures that there is no hyperplane that divides clean and triggered samples for any class. To accomplish this, we set $x^{(1)} := CDF_c^{-1}\big(1-CDF_c(x^{(1)})\big)$ where $CDF_c$ is the marginal CDF of pixel $x^{(1)}$ for class $class(x)$.

\begin{figure}[!htbp]
  \centering
    \includegraphics[width=.45\textwidth]{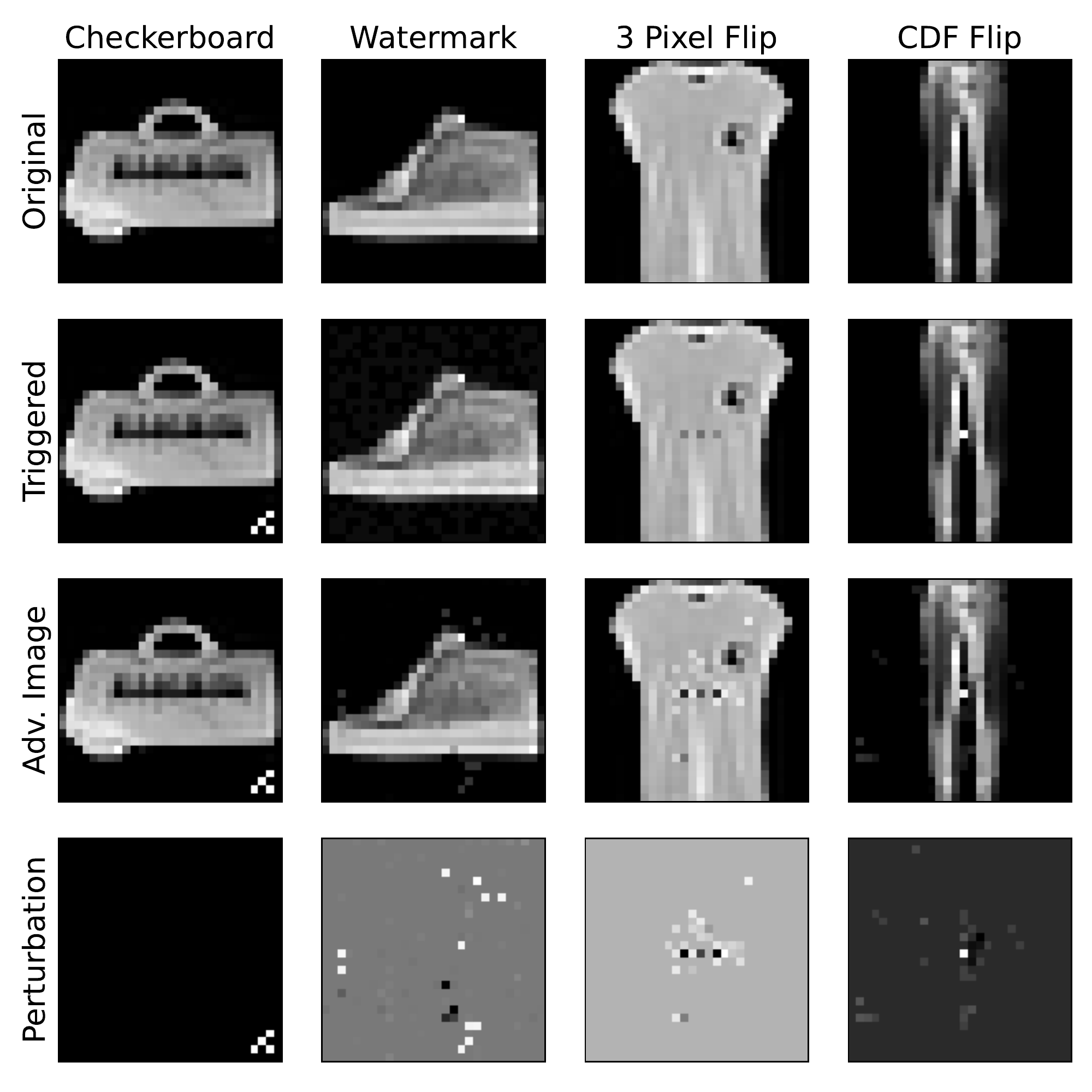}
    \caption{Fashion MNIST examples (top row) with different triggers applied (second row). Bottom two rows show images and perturbations resulting from reverse engineering the triggers from poisoned models.}\label{fig:shoe} 
\end{figure}
Figure \ref{fig:shoe} shows examples of all four triggers applied to the Fashion MNIST dataset. To verify that these triggers are not linearly separable from clean data, we trained logistic regression models poisoned with each trigger on the Fashion MNIST dataset \cite{xiao2017online} and examined their performance, shown in Table \ref{tab:linmodel}. 
Linearly separable triggers are easy for these models to spot and do not affect performance on clean data. 
On the other hand, linear models cannot fully separate the non-linear triggers from clean data, and poisoning them with the non-linear triggers drops performance on clean data (shown in bold). 

\begin{table}[!htbp]
\centering
\caption{The logistic regression performance on different trigger types.}\label{tab:linmodel}
\vspace{2mm}
\begin{tabular}{|c|c|c|}\hline
    \multirow{2}{*}{Trigger} & Clean & Trigger \\
     &  Accuracy & Accuracy \\\hline\hline
	None & 0.84 & --   \\\hline
    Checkerboard & 0.84 & 1.00 \\\hline
    Watermark & 0.84 & 1.00 \\\hline
    3 Pixel Flip & \textbf{0.67} & \textbf{0.66} \\\hline
    CDF Flip & \textbf{0.77} & \textbf{0.20} \\\hline
\end{tabular}
\end{table}

We trained two types of nonlinear models on Fashion MNIST with each of these triggers embedded in them. We used a shallow 4 layer convolutional neural network (CNN) and a 3 layer fully connected neural network (FCNN). 
Unlike logistic regression, the nonlinear models were able to fit both clean and triggered data relatively well. 
We generated adversarial perturbations on these models using a sparse $\ell_1$ PGD attack, implemented in AdverTorch \cite{ding2019advertorch}.  
We set the sparsity to 0.99 and used attack strength $\varepsilon = 5$.
We used 50 random starts of 10 step PGD and selected the perturbation with the maximal loss. 
Example perturbations and the resulting adversarial images are shown in the bottom rows of Figure \ref{fig:shoe}.
Using these perturbations, we computed our TOP metrics FR and FC, which we show in Table \ref{tab:trans}. 
Poisoning the models with any of these triggers results in elevated FR and FC metrics, regardless of whether the trigger is linearly separable from clean data or not. 
Applying a threshold to either metric leads to perfect separation of clean (in bold) and poisoned models. 


\begin{table}[!htbp]
\centering
\caption{TOP metrics on Fashion MNIST models.}
\label{tab:trans}
\vspace{2mm}
\resizebox{.98\linewidth}{!}{
\begin{tabular}{|c|c|c|c|c|c|}\hline
    \multirow{2}{*}{Model} & \multirow{2}{*}{Trigger} & Clean & Trigger & \multirow{2}{*}{FR} & \multirow{2}{*}{FC} \\
     & &  Accuracy & Accuracy & &\\\hline\hline
	\multirow{5}{*}{CNN} & None & 0.92 & -- & \textbf{0.08} & \textbf{0.02}\\\cline{2-6}
	& Checkerboard & 0.91 & 1.00  & 0.30 & 0.15\\\cline{2-6}
    & Watermark & 0.91 & 1.00 & 0.31 & 0.31 \\\cline{2-6}
    & 3 Pixel Flip & 0.91 & 0.93  & 0.79 & 0.77\\\cline{2-6}
    & CDF Flip & 0.90 & 0.73  & 0.81 & 0.75\\\hline
	\multirow{5}{*}{FCNN} & None & 0.88 & --  & \textbf{0.08} & \textbf{0.03}\\\cline{2-6}
    & Checkerboard & 0.88 & 0.99  & 0.50 & 0.34 \\\cline{2-6}
	& Watermark & 0.87 & 1.00 & 0.68 & 0.66 \\\cline{2-6}
	& 3 Pixel Flip & 0.86 & 0.87  & 0.64 & 0.60 \\\cline{2-6}
	& CDF Flip & 0.84 & 0.70  & 0.67 & 0.62 \\\hline
\end{tabular}
}
\end{table}




\section{Reverse Engineering Triggers}\label{section:reverse_eng_trigger}
\noindent The TOP method does not require pristine reconstructions of a Trojan trigger to work. 
However we have found in practice that good reverse engineering of the trigger tends to increase the separation of clean and poisoned models with respect to our TOP metrics. 
Good reverse engineering also offers concrete examples of what perturbations seem to be transferable across images, which may aid a human defender in reducing false detections or determining the nature of the true trigger.

Our general approach to reverse engineering triggers is to use PGD \cite{Madry2018TowardsDL} to solve for $\delta$, as defined in Equation \ref{eq:delta}, over a set of plausible triggers that form the attack domain $\Delta$. 
For ideal reverse engineering, the set $\Delta$ would be as small as possible while still containing the perturbations induced by the trigger function.  
Since the trigger function doesn't change the true class but does change the poisoned model's prediction, trigger perturbations generally produce large loss values relative to other perturbations and are good solutions to the maximization problem used in the definition of $\delta$ given on Equation \ref{eq:delta}.

Common triggers used in the literature include localized triggers (e.g., stickers, patches), watermarks, and filters (e.g., Instagram filters). 
We define $\Delta$'s that approximately correspond to these different types of triggers.  
A localized trigger generally induces large changes to a small set of pixels.  
An attack domain based on a bound on the $\ell_0$ norm or a sparse $\ell_1$ norm bound encompasses such triggers. 
For watermarks, $\ell_{\infty}$ or $\ell_{2}$ bounded attack domains roughly capture possible triggers. 
\subsection{Adversarial Filters}
\noindent Filter triggers typically modify an image by a large amount in terms of any $\ell_p$ norm, so traditional adversarial attack domains are not particularly well suited reverse engineering this type of trigger.
To address this case, we introduce the \textit{adversarial filter}.  
Instead of constraining the norm of the additive perturbation, we constrain the norm of a convolutional filter.  
Formally, we solve the following optimization problem for filter $w$:
\begin{align} \label{equ:untargeted}
    \max_{w \in \cW} \cL \Big(f(x), f(x + w*x)\Big) 
\end{align}
where the search space $\cW = \big\{w\, \big|\, ||w|| \leq \varepsilon \big\}$ for some norm $||\cdot||$. 
We have found $\ell_{\infty}$ and $\ell_{2}$ to serve effectively as the norm.
Adversarial filters give us a set of perturbations with a very small measure in image space, but large constraints on perturbation norm, mirroring the behavior of Instagram triggers.
Our experiments show that this technique gives a significant improvement in detecting poisoned models with Instagram triggers. 
\subsection{Combining Attack Domains}

\noindent Factors such as adversarial training, network architecture, and trigger type can strongly affect the response of a neural network to a particular adversarial attack.  
To make the TOP algorithm as robust as possible to these different factors, we compute FR and FC separately on a variety of adversarial perturbation strengths and types.  
We use logarithmically spaced attack magnitudes, spanning from very small attacks with little impact on the classifiers to very large attacks that saturate the FR and FC metrics.
We then treat these as features for a simple classifier that can pick up TOP signals in a variety of models. 
We use a logistic regression classifier and an iterative feature selection procedure to combine these scores into a final probability of poisoning. 
Given some set of training models, we train a logistic regression classifier on all features. 
We then prune features with negative weights and repeat this procedure until all weights are positive.
Since we expect all TOP metrics to be positively correlated with poisoning, this is a way of incorporating our prior beliefs as a regularizer on the training process. 
We found that this procedure consistently improves top level metrics, especially with small training sets.





\section{Experiments}\label{section:experiments}
\noindent In this section, we perform experiments to evaluate our TOP method and show that our method does not require extensive tuning to achieve reasonable performance.

\subsection{Datasets}
\noindent\textbf{Fashion MNIST.} We trained 40 models of various architectures on the Fashion MNIST dataset \cite{xiao2017online}.  Half of these models were poisoned with the 4 triggers discussed in section \ref{section:transferability}. 
We randomly selected the architecture from a set consisting of fully connected networks with \{3, 4\} layers or CNNs with \{5, 6, 7, 8, 9\} layers. 
For the poisoned models, we randomly selected one of the 4 triggers and a target class.  
We trained models for 30 epochs with the Adam optimizer \cite{Kingma2015AdamAM}. We ensured that models met minimum a classification threshold of 0.6 on clean and triggered data.

\begin{figure}[H]
  \centering
    \includegraphics[width=.325\textwidth]{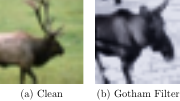} 
    \caption{Examples of both clean and triggered images from the CIFAR-10 dataset.}\label{fig:cifar10_examples} 
\end{figure}

\noindent\textbf{CIFAR-10.} We also trained 10 models on the CIFAR-10 dataset \cite{cifar10Alex}, a well-known image classification dataset with an image size of 32x32. 
The models have different network architectures consisting of DenseNet-\{121, 161, 169, 201\}. Half of the models are clean and the other half of the models are poisoned with Instagram Gotham Filter attack using the TrojAI software framework \cite{karra2020trojai}, an open source set of Python tools capable of generating triggered (poisoned) datasets and associated deep learning models with trojans. Figure \ref{fig:cifar10_examples} shows an example of a clean example and an example of a Gotham filter-based adversarial example from the CIFAR-10 dataset. 

\begin{figure}[H]
  \centering
    \includegraphics[width=.45\textwidth]{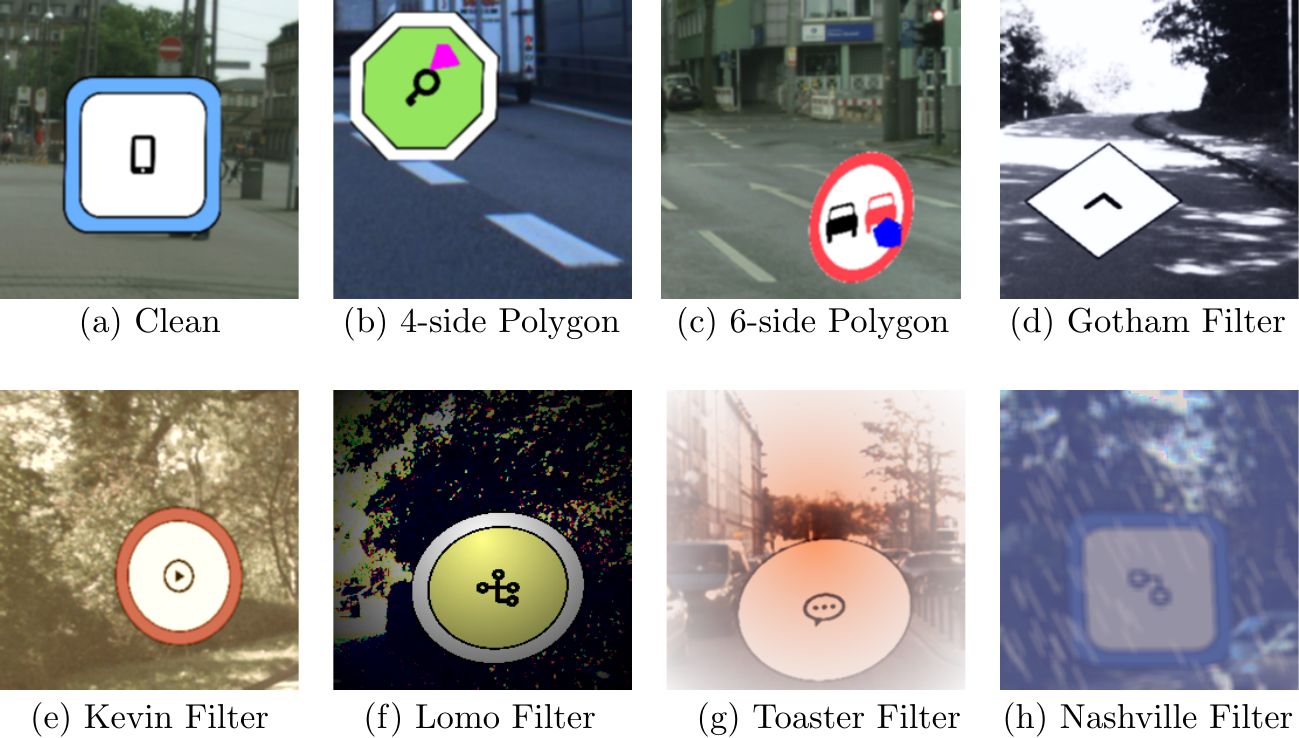} 
    \caption{Examples of both clean and triggered images from the TrojAI benchmark dataset.}\label{fig:trojai_examples} 
\end{figure}

\noindent\textbf{TrojAI benchmark.} The TrojAI program, organized by IARPA \cite{IARPA_trojai_challenge}, aims to tackle the backdoor detection problem by defining a set of public benchmarks, presented in ``rounds". Each round defines a set of clean and poisoned training models. The benchmark task is to predict whether models in a test set are clean or poisoned. We report results on TrojAI rounds 1-3. These rounds involve models that are trained on synthetically generated traffic sign data. 
Round 1 uses randomly generated polygons as triggers. 
A polygon trigger has a randomly chosen number of sides between 3 and 12 and a color chosen uniformly at random from $[0,1]^3$.
Rounds 2 and 3 use both polygons and five specific Instagram filters as triggers.
Figure \ref{fig:trojai_examples} depicts examples of clean and triggered images from the TrojAI benchmark dataset. 

We briefly introduce a few notations to describe the dataset. 
Let $M$ and $M'$ be sets of neural network models for training and testing.  Let $G$ be a set of neural network architectures, $C$ be a set of class labels, and $D$ be a set of validation images. Table \ref{tab:trojai_dataset} summarizes the statistics of the three rounds.  Round 1 uses DenseNet121,  InceptionV3, ResNet50 architectures. 
Rounds 2 and 3 use various network architectures.

\begin{table}[!htbp]
\centering
\caption{The summary of TrojAI benchmark datasets.}
\label{tab:trojai_dataset}
\vspace{2mm}
\begin{tabular}{|c|c|c|c|c|c|}\hline
     Dataset & $|M|$ & $|M'|$ & $|G|$ & $|C|$ & $|D|$ \\\hline\hline
     Round 1 & 1000 & 100 &3  &5  & $100|C|$ \\\hline
     Round 2 & 1104 & 144 &22 & $\leq 25$  & $\leq 20|C|$\\\hline
     Round 3 & 1108 & 288 &22 & $\leq 25$ & $\leq 20|C|$\\\hline
    \end{tabular}
\end{table}



\subsection{Evaluation Metrics}
\noindent We use two accuracy metrics consistent with the metrics used in TrojAI competition \cite{IARPA_trojai_challenge}: Area under Receiver Operating Characteristic Curve (AUC)  \cite{BradleyROCAUC1997} and cross-entropy loss (CE) \cite{MannorCE2005}. 
AUC captures how well a classifier separates clean and poisoned models by integrating detection and false alarm probabilities at different thresholds.
CE captures both class separation \textit{and} how well calibrated the predicted probability of poisoning is.  CE is a more stringent metric. 
AUC is in the $[0,1]$ interval where perfect separation gives a score of 1.0 and random guessing gives an AUC of 0.5.
CE is in the $[0,\infty)$ interval where perfect, confident classification gives a CE of 0 and assigning a probably of 0.5 to all samples gives a CE of $ln(2) \approx 0.693$.  
The TrojAI program sets a CE score of $ln(2)/2 \approx 0.347$ as a detection performance goal.

\subsection{Experimental Results}
\noindent\textbf{Results for Fashion MNIST.} We used the sparse $\ell_1$ attack outlined in section \ref{section:transferability} with sparsity $=0.99$ and $\varepsilon = 5$. We note that while this attack is well-suited to some of the triggers, it is not particularly well-suited to the watermark trigger, which is not sparse.  
We used these perturbations to compute FC. Since we only use one detection score here, we can compute AUC without any additional scaling. We use Platt scaling \cite{Platt1999ProbabilisticOF} (equivalent to univariate logistic regression) to arrive at a calibrated probability based on a small set of ``training" models $M$.
We randomly sample class-balanced calibration sets of different sizes and perform 500 evaluations. Table \ref{tab:fmnist1} shows the results of this process. 
We ran an additional experiment in which we train our detector on one type of trigger and evaluate it on other types. These results are shown in Table \ref{tab:fmnist2}.
These results show that TOP provides accurate and well calibrated detection probabilities with just a few training models and can be effective in detecting novel triggers.

\begin{table}[!htbp]
\centering
\caption{Fashion MNIST results by training set size.}
\label{tab:fmnist1}
\vspace{2mm}
\begin{tabular}{|c|c|c|}\hline
     $|M|$ & AUC & CE \\\hline\hline
     1 & 0.962 & 0.382  \\\hline
     2 & 0.962 & 0.341  \\\hline
     3 & 0.962 & 0.322  \\\hline
     5 & 0.962 & 0.307  \\\hline
     10 & 0.962 & 0.275 \\\hline
    \end{tabular}
\end{table}


\begin{table}[!htbp]
\centering
\caption{Fashion MNIST CE results by trigger type.}
\label{tab:fmnist2}
\vspace{2mm}
\resizebox{1\linewidth}{!}{
\begin{tabular}{|c|c|c|c|c|}\hline
     \diagbox{Training}{Testing} & Checker & Watermark & 3 Pixel & CDF \\\hline\hline
     Checker & -- & 0.329 & 0.181 & 0.239 \\\hline
     Watermark & 0.370 & -- & 0.258 & 0.321 \\\hline
     3 Pixel& 0.430 & 0.362 & -- & 0.027 \\\hline
     CDF & 0.335 & 0.265 & 0.072 & --  \\\hline
    \end{tabular}
}
\end{table}

\begin{figure*}[!htbp]
\centering
\subfigure[Round 1]{%
\label{fig:r1fr}%
\includegraphics[width=.33\textwidth]{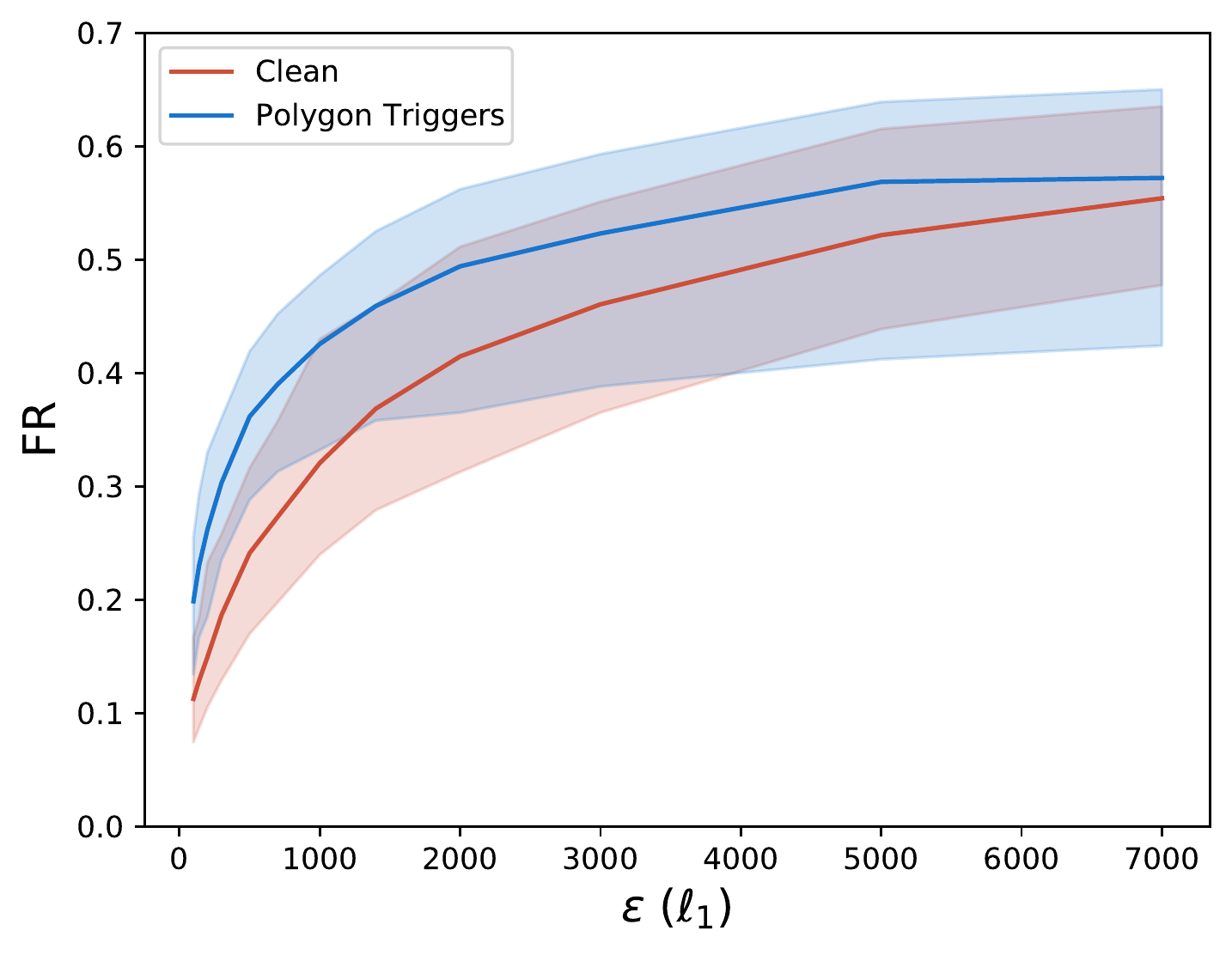}}%
\subfigure[Round 2 ]{%
\label{fig:r2fr}%
\includegraphics[width=.33\textwidth]{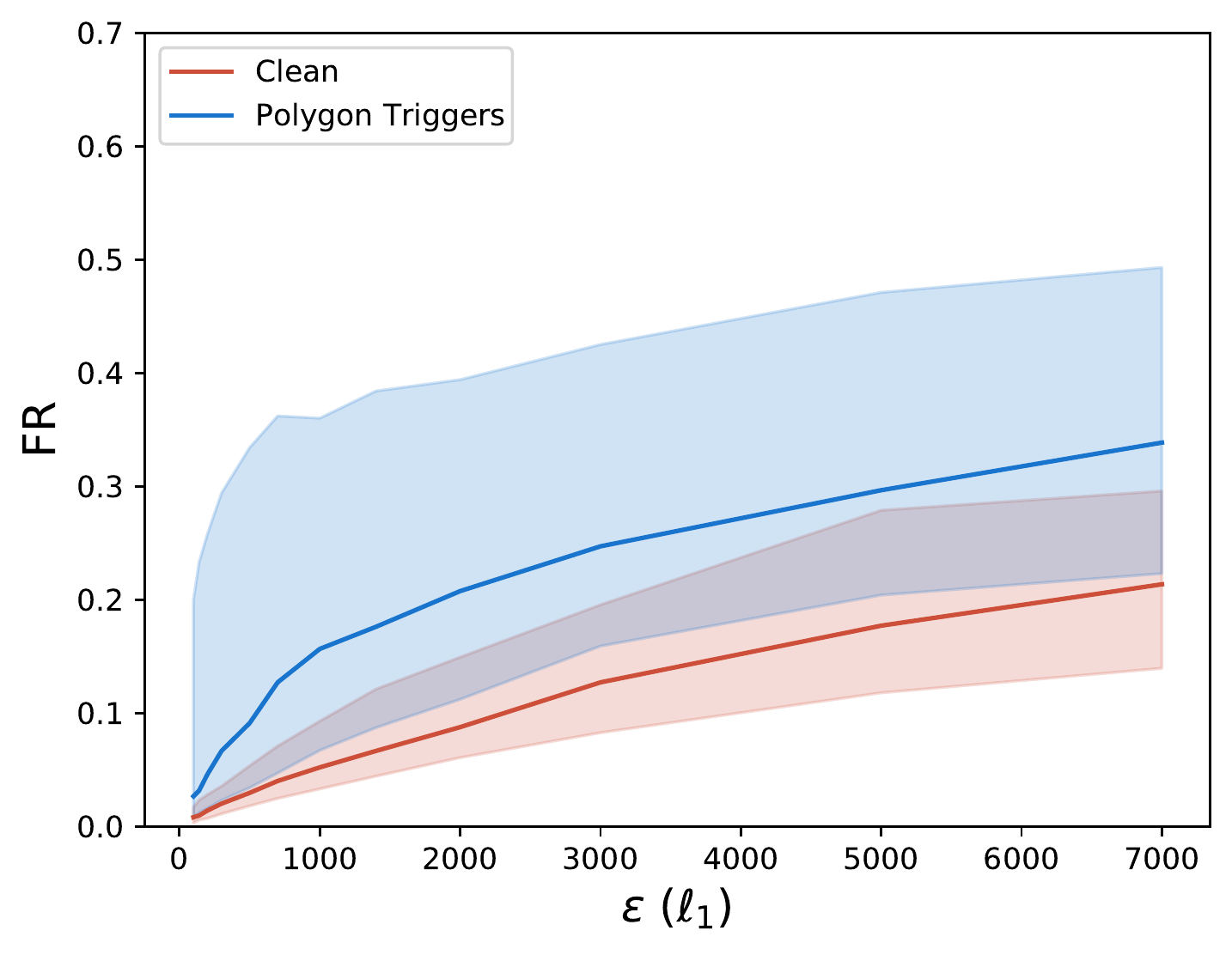}}%
\subfigure[Round 3 ]{%
\label{fig:r3fr}%
\includegraphics[width=.33\textwidth]{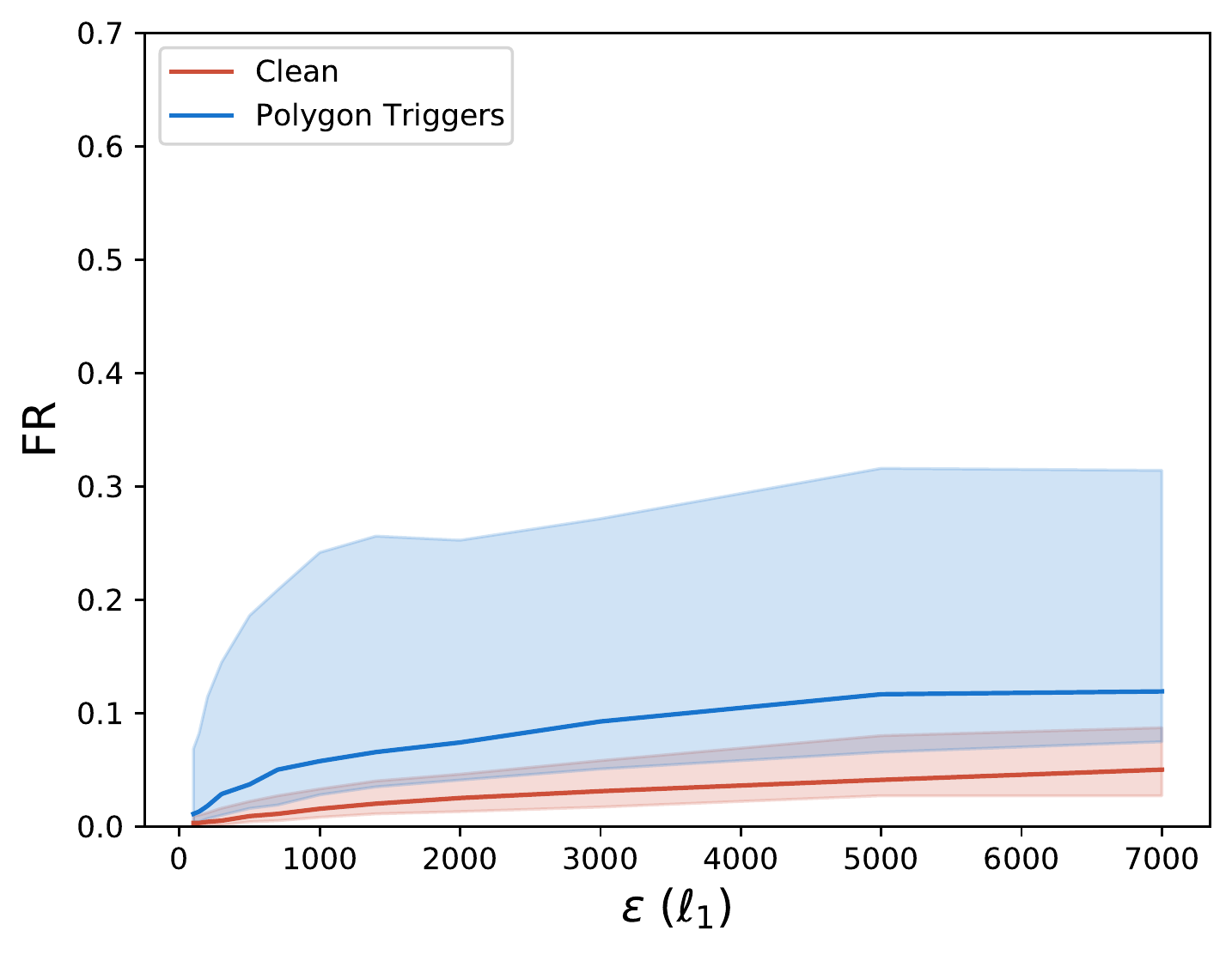}}%
\caption{Median and 80\% confidence bounds on FR scores for TrojAI benchmark models at different $\ell_1$ attack strengths.}\label{fig:fr_det0}
\end{figure*}

\begin{figure*}[!htbp]
\centering
\subfigure[Round 1]{%
\label{fig:r1fc}%
\includegraphics[width=.33\textwidth]{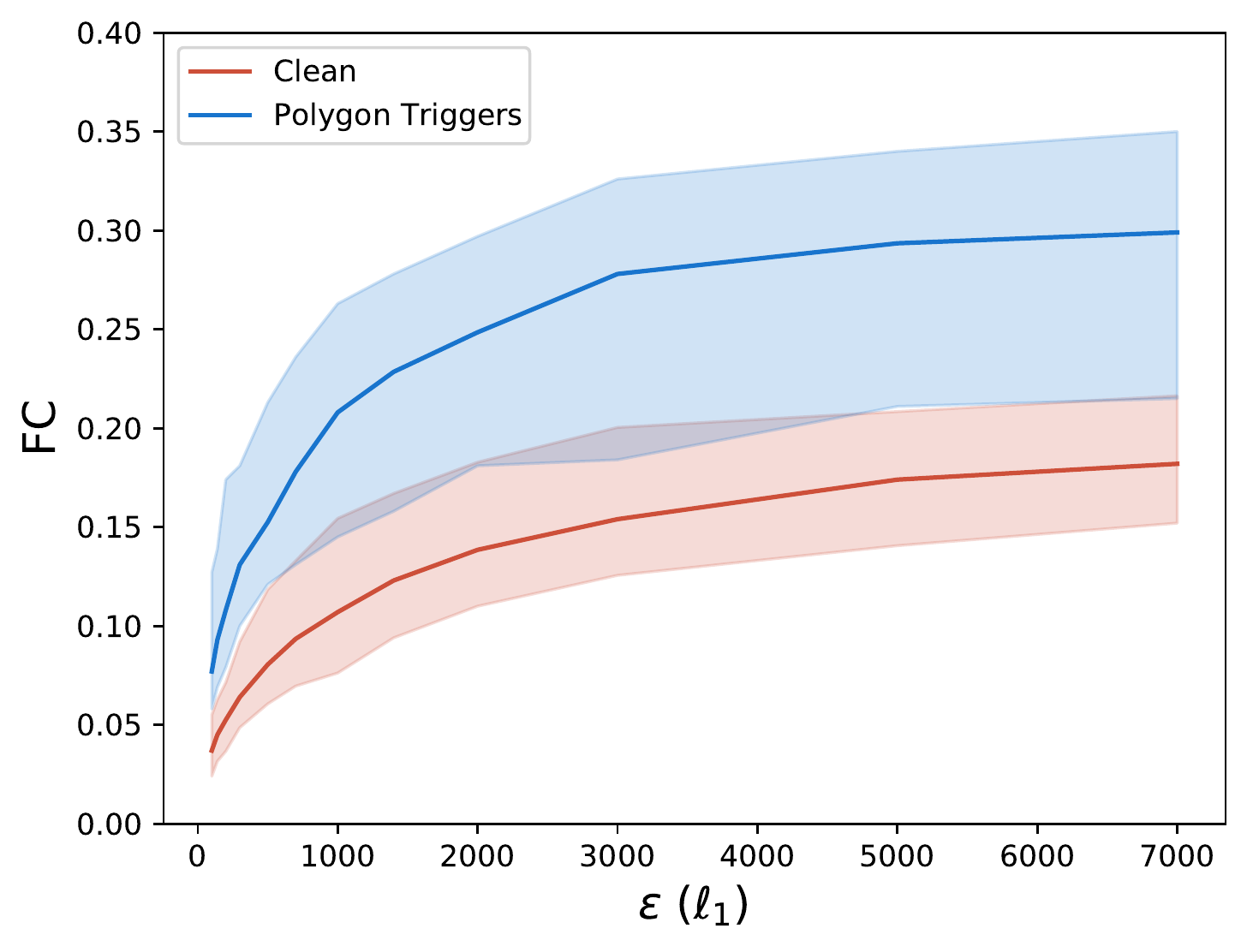}}%
\subfigure[Round 2]{%
\label{fig:r2fc}%
\includegraphics[width=.33\textwidth]{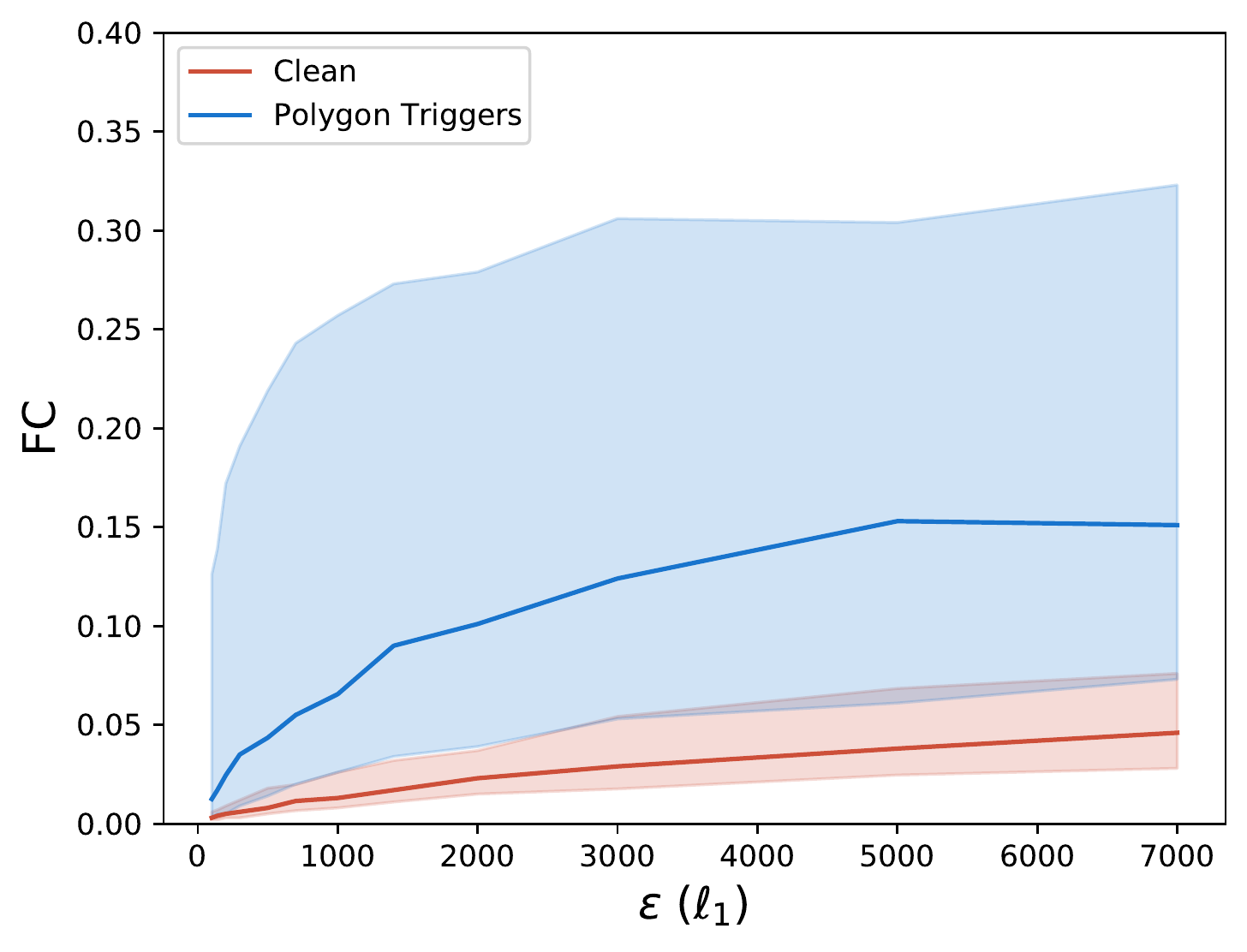}}%
\subfigure[Round 3]{%
\label{fig:r3fc}%
\includegraphics[width=.33\textwidth]{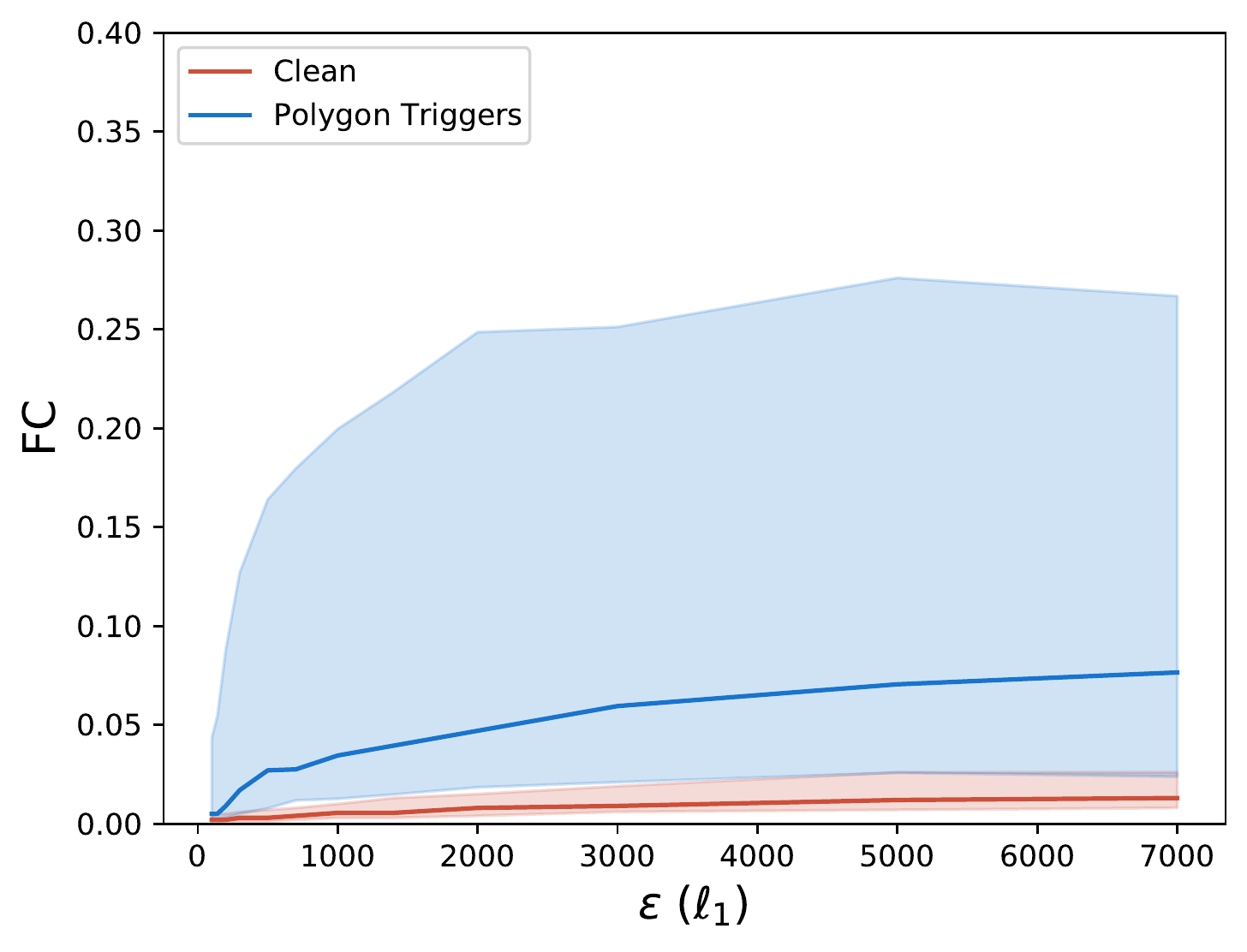}}%
\caption{Median and 80\% confidence bounds on FC scores for TrojAI benchmark models at different $\ell_1$ attack strengths.}\label{fig:fc_det0}
\end{figure*}

\noindent\textbf{Results for TrojAI.} 
We start by comparing our two metrics, fool rate and fool concentration, on the TrojAI dataset. 
Figures \ref{fig:fr_det0} and \ref{fig:fc_det0} compare the effectiveness of FR and FC in discriminating between clean and poisoned models at different attack magnitudes across different rounds. 
Figure \ref{fig:fr_det0} shows the median and 80th percentile spread of FR scores for all attack strengths for rounds 1, 2, and 3 respectively. Figure \ref{fig:fc_det0} shows the same for FC. 
While fool rate is able to discriminate well in all three rounds, fool concentration provides a stronger signal. 
We use FC as the basis for the subsequent results in this section.



We calibrated our detector on a random partition of 25\% of the training set provided by the TrojAI benchmark. 
Table \ref{tab:trojaires} shows the top line metrics for different subsets of the rounds and triggers.





\begin{table}[!htbp]
\centering
\caption{The top-level results on TrojAI benchmark models.}
\label{tab:trojaires}
\vspace{2mm}
\begin{tabular}{|c|c|c|c|}\hline
Models & Trigger(s) & CE & AUC \\\hline\hline
Round 1 & Polygon & 0.42 & 0.87 \\\hline
Round 2 & Polygon & 0.40 & 0.89 \\\hline
Round 2 & Polygon+Instagram & 0.49 & 0.85 \\\hline
Round 2 & Instagram & 0.44 & 0.86 \\\hline
Round 3 & Polygon & 0.35 & 0.90 \\\hline
Round 3 & Instagram & 0.49 & 0.79 \\\hline
Round 3 & Polygon+Instagram & 0.50 & 0.83 \\\hline
\end{tabular}
\end{table}



\begin{figure*}[!htbp]
\centering
\subfigure[AUC]{%
\label{fig:r123auc}%
\includegraphics[width=.4\textwidth]{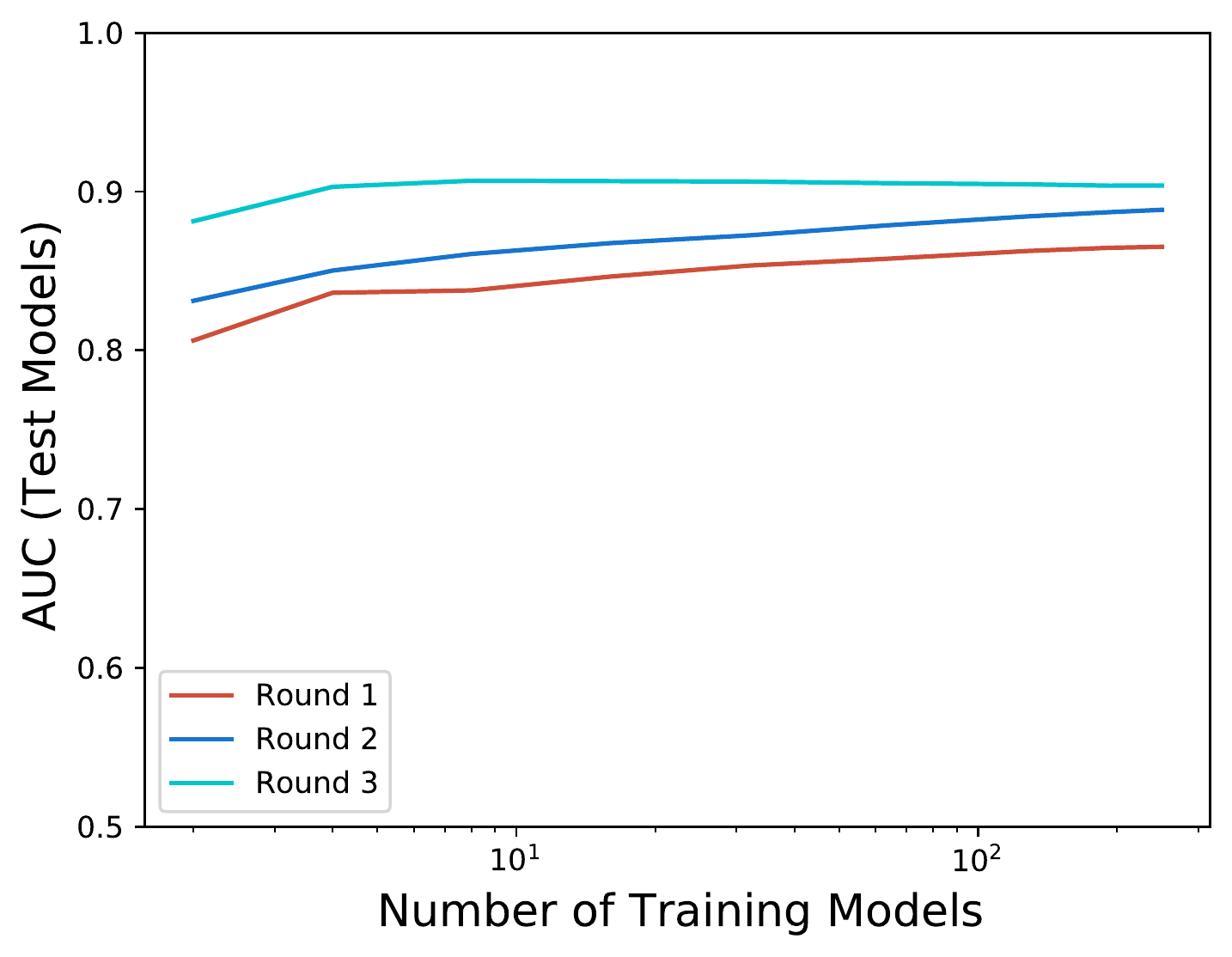}}%
\subfigure[Cross Entropy]{%
\label{fig:r123ce}%
\includegraphics[width=.4\textwidth]{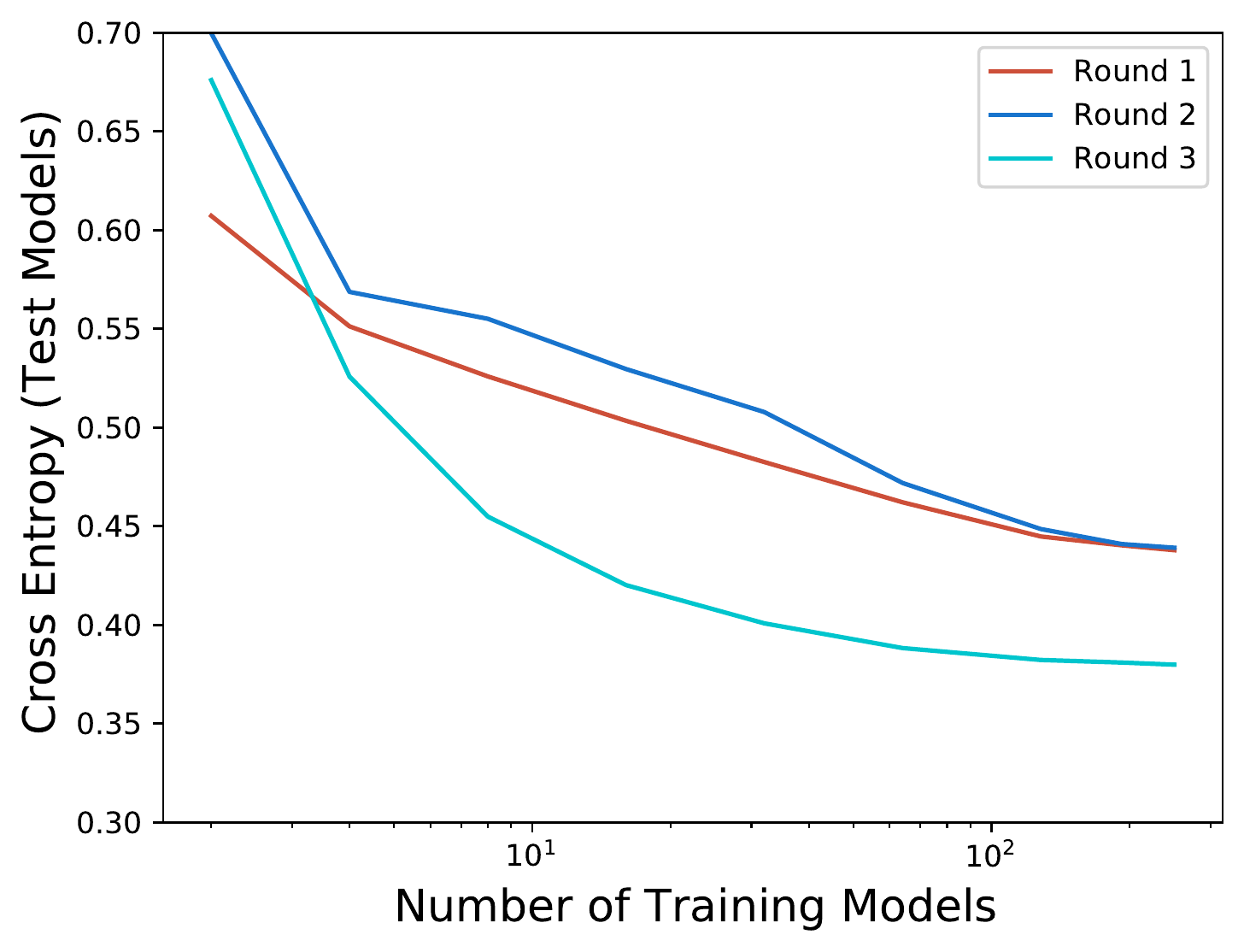}}%
\caption{The performance of TOP on Clean/Polygon models as a function of the training set size.}
\end{figure*}

We also examined our detector's performance when it is trained on small training sets. 
We randomly sampled class-balanced subsets of the rounds 1-3 training sets, calibrated the detector, and evaluated it on the test models from the respective round.
We performed this experiment 200 times for each training set size.  
Figures \ref{fig:r123auc} and \ref{fig:r123ce} show how the number of training models impacts the top level metrics on the test models. 
Even with just a single positive and negative example, our detector can effectively separate clean and poisoned models, with an AUC over 0.8 for all rounds. 
Cross-entropy is a more challenging metric, but our detector can achieve the 0.5 level with between 4 and 32 models, depending on the round.  

 \begin{figure*}[!htbp]
\centering
\subfigure[Round 2]{%
\label{fig:rnd2_attack_auc}%
\includegraphics[width=.4\textwidth]{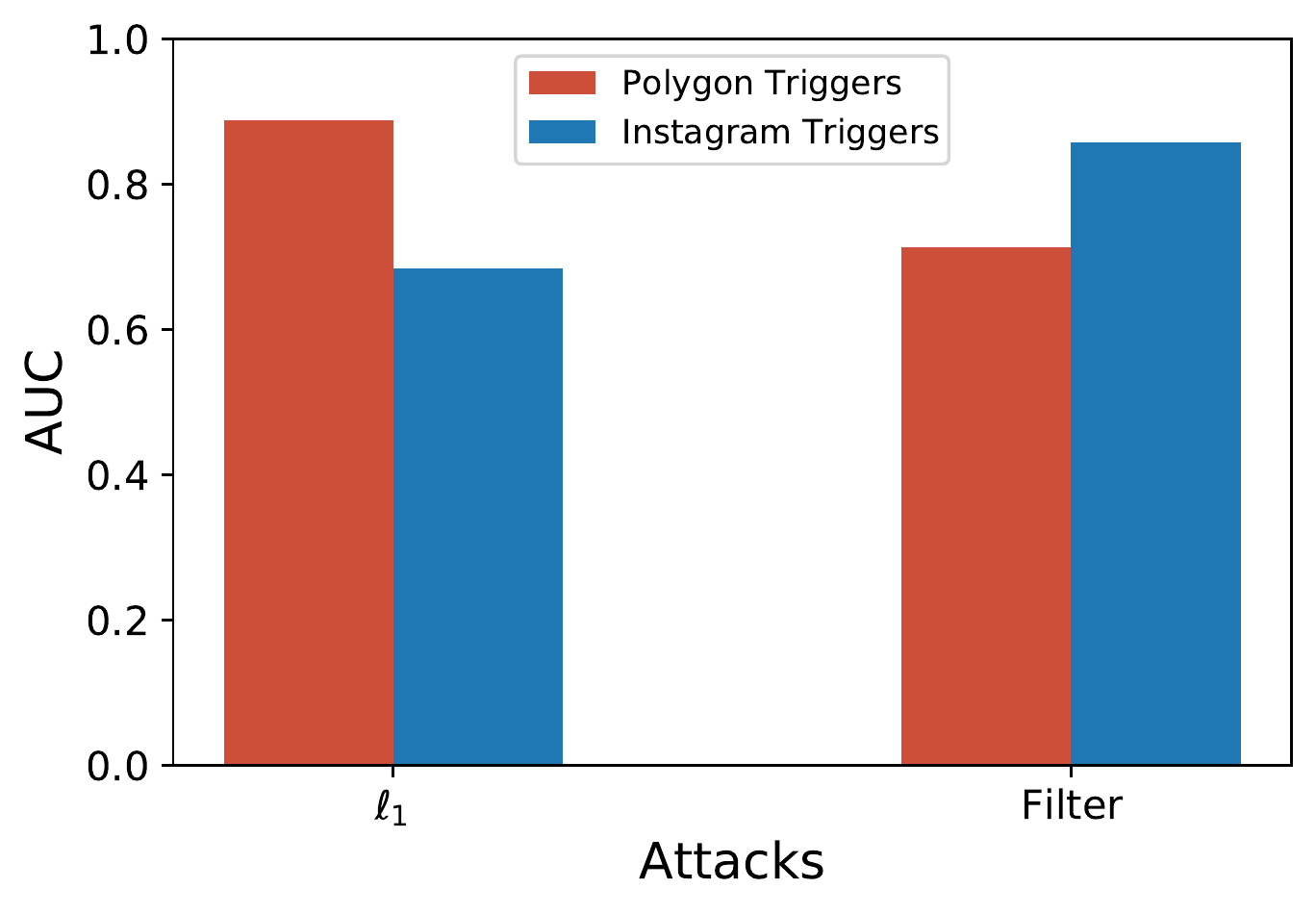}}%
\subfigure[Round 3]{%
\label{fig:rnd3_attack_auc}%
\includegraphics[width=.4\textwidth]{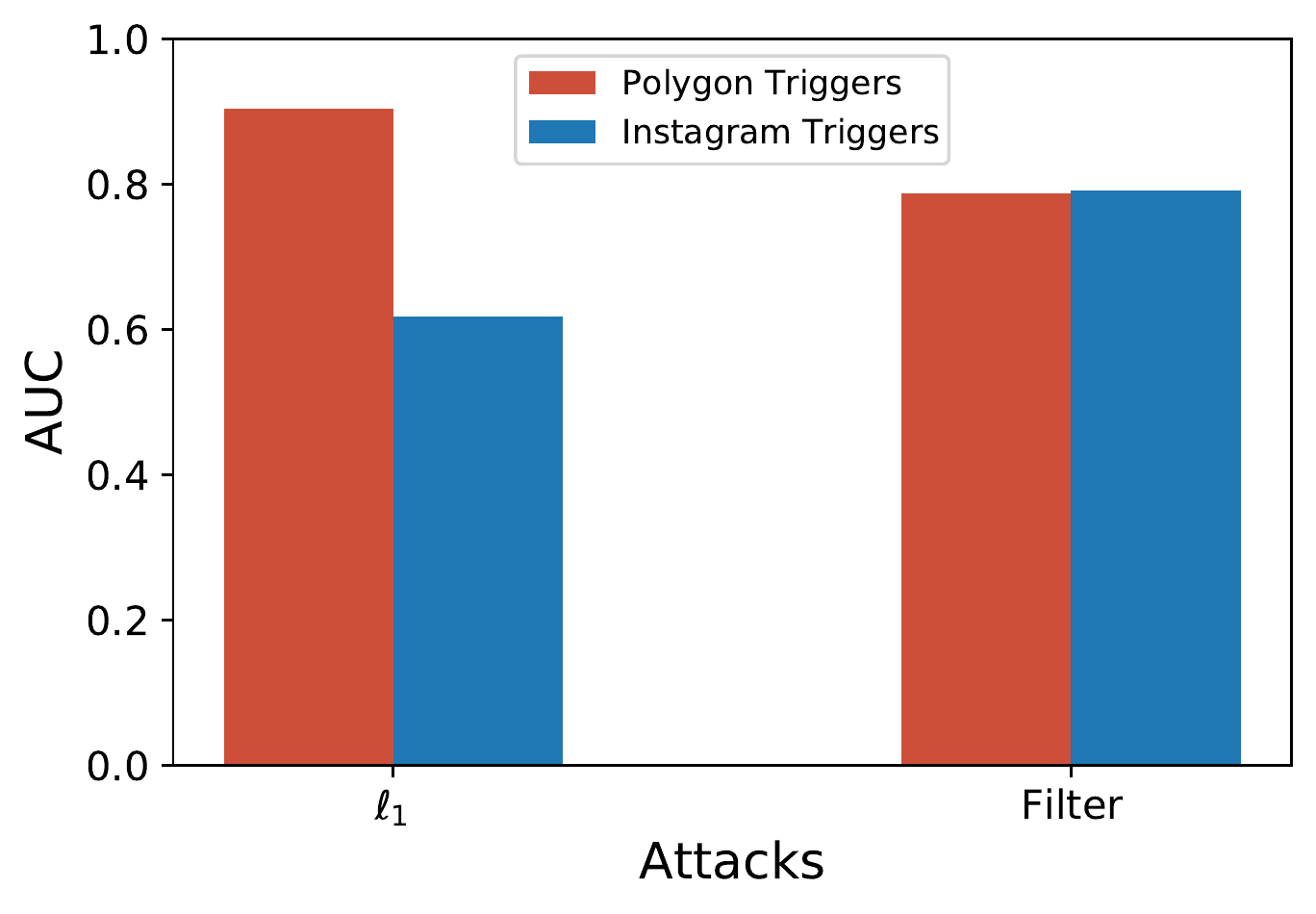}}%
\caption{Detecting two types of triggers with different attack domains. }\label{fig:cross_attack}
\end{figure*}

We now look at how well our detector functions without independent and identically distributed (IID) data to train on. This is a very challenging setting, but a very important one for practical backdoor detection.  To test this, we train our detector on each round, then test it on all three rounds.  
This forms a matrix of AUC scores, shown in table \ref{tab:crossround_all}, in which the diagonal values represent IID performance and off-diagonal represents non-IID performance.  
While the performance is far from perfect, TOP is able to discriminate between clean and poisoned models in a non-IID setting with an AUC around 0.75. 


  
\begin{table}[!htbp]
\centering
\caption{The cross-round AUC performance  for all models in the TrojAI benchmark datasets.}
\label{tab:crossround_all}
\vspace{2mm}
\begin{tabular}{|c|c|c|c|}\hline
\diagbox{Training}{Testing} & Round1 & Round2 & Round3 \\ \hline \hline
Round1 & \textbf{0.87}   & 0.77   & 0.70  \\ \hline
Round2 & 0.88   & \textbf{0.78}   & 0.75  \\ \hline
Round3 & 0.68   & 0.79   & \textbf{0.76}  \\ \hline
\end{tabular}
\end{table}
  

Figure \ref{fig:cross_attack} shows how well TOP works with ``matched" and ``mismatched" attacks. 
Even though we designed adversarial filters with Instagram triggers in mind, they still are fairly effective serving as the basis for detecting models with polygon triggers, providing AUC scores above 0.7 in both rounds. 
Sparse $\ell_1$ attacks are not as effective on Instagram triggers, but they still provide a signal that is better than guessing.

  

\noindent\textbf{Results for CIFAR10.} To show the robustness of TOP in different settings, we used the detector from our TrojAI experiments on the CIFAR10 models. 
We compute our standard set of attacks and calibrate on 8 models. 
We also calibrated a detector using only the filter-based attacks. 
We used repeated random subsetting in which we trained on 8 models and tested on 2 to allow us to compute mean CE and AUC metrics over multiple trials.
Table \ref{tab:cifar10} shows results for a TOP detector based on only filter attacks, as well as the full $\ell_1$ and filter attack set. 
Since the true trigger is a filter, our filter attacks work especially well on it. 
However, the full detector works nearly as well with minimal training.
\begin{table}[H]
\centering
\caption{The top-level results on CIFAR10 models.}
\label{tab:cifar10}
\vspace{2mm}
\begin{tabular}{|c|c|c|c|}\hline
    Attacks & CE & AUC \\\hline\hline
	Filter & 0.46 & 0.92 \\\hline
    Filter + $\ell_1$ & 0.54 & 0.84 \\\hline
\end{tabular}
\end{table}

\section{Conclusion and Future Work}
\noindent In this paper, we identified an interesting property of trained deep neural network models - that adversarial perturbations transfer from image to image more readily in poisoned models than in clean models. 
We showed that this transferabilty property holds for a variety of model and trigger types, including triggers that are not linearly separable from clean data.  
We used this feature to detect poisoned models in the TrojAI benchmark, as well as other dataset.
We showed that TOP is a robust indicator of backdoor poisoning, even in challenging non-IID settings, and in settings without many example models.


\section*{Acknowledgments}

\noindent This effort was supported by the Intelligence Advanced Research Projects Agency (IARPA) under the contract W911NF20C0034. The content of this paper does not necessarily reflect the position or the policy of the Government, and no official endorsement should be inferred.

We would like to thank Jeremy E.J. Cohen for his thoughtful discussions and feedback.

{\small
\bibliographystyle{ieee_fullname}
\bibliography{egbib}
}

\end{document}